\documentclass[10pt,twocolumn,letterpaper]{article}

\usepackage{cvpr}              

\usepackage{graphicx}
\usepackage{amsmath}
\usepackage{amssymb}
\usepackage{booktabs}

\usepackage[pagebackref,breaklinks,colorlinks]{hyperref}

\usepackage[capitalize]{cleveref}
\crefname{section}{Sec.}{Secs.}
\Crefname{section}{Section}{Sections}
\Crefname{table}{Table}{Tables}
\crefname{table}{Tab.}{Tabs.}

\def\cvprPaperID{71} 
\def\confName{CVPR}
\def\confYear{2022}

\begin{document}

\title{IMDeception: Grouped Information Distilling Super-Resolution Network}

\author{Mustafa Ayazoğlu\\
Aselsan Research\\
Ankara, Turkey\\
{\tt\small mayazoglu@aselsan.com.tr}
}
\maketitle

\begin{abstract}
Single-Image-Super-Resolution (SISR) is a classical computer vision problem that has benefited from the recent advancements in deep learning methods, especially the advancements of convolutional neural networks (CNN). Although state-of-the-art methods improve the performance of SISR on several datasets, direct application of these networks for practical use is still an issue due to heavy computational load. For this purpose, recently, researchers have focused on more efficient and high-performing network structures. Information multi-distilling network (IMDN) is one of the highly efficient SISR networks with high performance and low computational load. IMDN achieves this efficiency with various mechanisms such as Intermediate Information Collection (IIC), working in a global setting, Progressive Refinement Module (PRM), and Contrast Aware Channel Attention (CCA), employed in a local setting. These mechanisms, however, do not equally contribute to the efficiency and performance of IMDN. In this work, we propose the Global Progressive Refinement Module (GPRM) as a less parameter-demanding alternative to the IIC module for feature aggregation. To further decrease the number of parameters and floating point operations per second (FLOPS), we also propose Grouped Information Distilling Blocks (GIDB). Using the proposed structures, we design an efficient SISR network called IMDeception. Experiments reveal that the proposed network performs on par with state-of-the-art models despite having a limited number of parameters and FLOPS. Furthermore, using grouped convolutions as a building block of GIDB increases room for further optimization during deployment. To show its potential, the proposed model was deployed on NVIDIA Jetson Xavier AGX and it has been shown that it can run in real-time on this edge device.

\end{abstract}

\section{Introduction}

\begin{figure}[h]
\centering
\subfloat[Original: Div2K 0886]{\includegraphics[trim={0 0 0 50},clip,width=1\linewidth]{"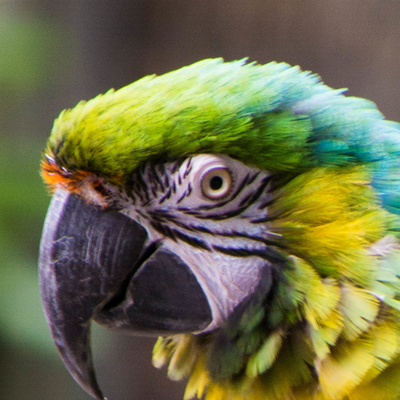"}} \\
\subfloat[\textbf{IMDeception\\(ours)}]{\includegraphics[trim={100 150 150 100},clip, width=0.25\linewidth]{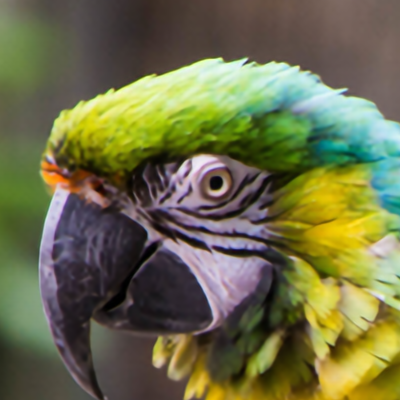}}
\subfloat[CARN]{\includegraphics[trim={100 150 150 100},clip,width=0.25\linewidth]{"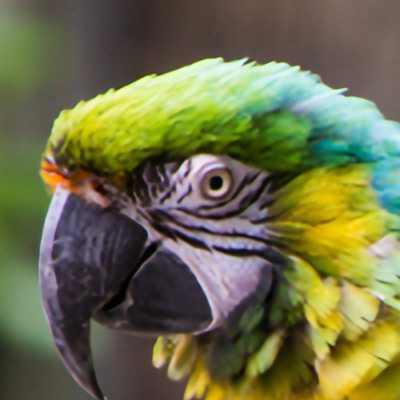"}}
\subfloat[IMDN]{\includegraphics[trim={100 150 150 100},clip,width=0.25\linewidth]{"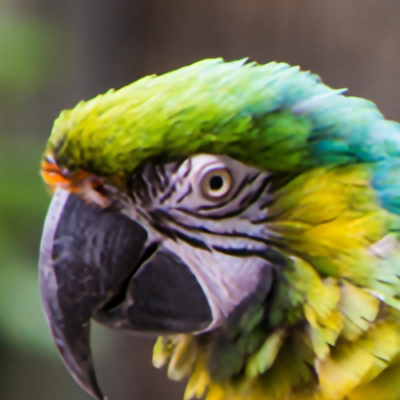"}}
\subfloat[Bicubic]{\includegraphics[trim={100 150 150 100},clip, width=0.25\linewidth]{"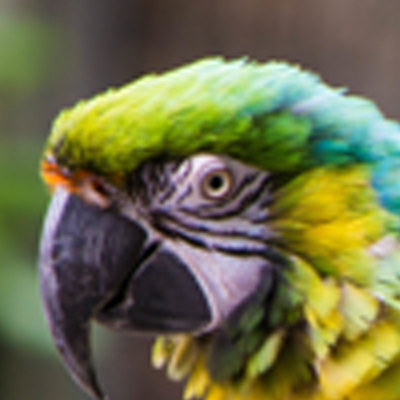"}}

   \caption{Results of our method compared with other methods}
\label{fig:900}
\end{figure}

\raggedbottom

Single-Image-Super-Resolution (SISR) is a well-studied computer vision problem. The problem's goal is to create a high-resolution image from a single low-resolution image. Due to its nature, it is an ill-posed problem. Starting with the seminal work of Dong \etal \cite{SRCNN} the problem is addressed by using deep-learning approaches. Dong's model used a CNN with only 3-layers and beat the traditional approaches. Later on, to decrease the computational load, FSRCNN \cite{FSRCNN} proposed postponing the upscaling to the end of the network while most of the computation and feature extracting done in low-resolution. Shi. \etal proposed ESPCN, \cite{ESPCN}  which replaced the transposed convolution layer with Depth2Space operator. Later, Kim \etal proposed \cite{VDSR}, a 20-layer network, and showed that increasing the number of parameters can improve the network's performance. EDSR proposed by Lim \etal \cite{EDSR} further improved the state-of-the-art by increasing the number of layers and omitting BatchNorm layers from the network. Later on, Yu \etal proposed WDSR \cite{WDSR}, a network with 75M parameters and improved super-resolution results. Indeed, increasing the number of parameters improves the performance of a network, but it also makes it harder to use it in many practical real-time scenarios. For these reasons, researches started working on efficient models which aim to maintain image reconstruction performance with those of millions-of-parameter-networks while still being applicable for real-time scenarios \cite{AIM2020,NTIRE22}. To decrease the number of parameters, recursive networks are employed \cite{DRCN,DRRN}, but the number of FLOPS is very high for these networks. Besides this work, there are some work incorporating the attention idea into the SISR domain, such as \cite{HAN, RCAN}, which increases the receptive field and hence the performance of the network while keeping the parameters low at the cost of an increased number of operations.

In this context, Hui \etal proposed IDN \cite{IDN} which uses channel splitting method to separate the high-level features from the low-level ones ﻿while keeping the number of parameters low and maintaining acceptable performance. IMDN \cite{IMDN} further investigated the channel splitting idea at a granularity level and further improved the performance and inference speed. Besides channel splitting, IMDN employed Intermediate Information Collection (IIC) at the global level to accumulate the information from different information multi-distilling blocks (IMDB) and in the IMDBs it used Progressive Refinement Module (PRM) which splits the outputs of different convolution layers such that a portion of the information is directly flows to the end of the block while the rest is fed to the next Conv2d layer for further refinement.

Although IMDN is an efficient and well-performing network, global information fusion modules (IIC) and IMDB blocks are not ideal and there is still room for improvement. To this end, following the Network-in-Network \cite{NIN} and Inception \cite{Inception} spirit, we propose the Global Progressive Refinement Module (GPRM), which is an extension of the PRM in the global setting, in-place of the IIC module. Using GPRM gives us the flexibility to control the number of parameters while being able to integrate the mid-level information to the end of the network. To further reduce the number of parameters and operations, we proposed grouped information distilling blocks (GIDB) as the building blocks that employ grouped convolutions. Using grouped convolutions increases the room for further optimization during deployment. Furthermore, by incorporating the block-based non-local attention (NLA) blocks at the global level, \cite{NLA} we further improved the performance of the proposed model.

Reconstruction efficiency of the model is shown in various different datasets, and inference efficiency is shown using NVIDIA TensorRT since it is training framework agnostic and optimizes the network for the hardware at the hand.

\section{Related Works}

As with many computer vision problems, SISR has benefited a lot from the recent advancements in deep learning. The first SISR model using deep learning started with the work of Dong \cite{SRCNN}. Later on, by postponing the upscaling stage to the end and processing the input image at a lower resolution, FSRCNN \cite{FSRCNN} improved the inference speed. FSRCNN also replaced ReLU activation with PReLU. Later on VDSR \cite{VDSR} introduced a deeper network and introduced a long upscaling skip connection. These showed that deeper networks improve the performance and long skip connection helps with the optimization. The same spirit continued with recursive architectures where a shared parameter sub-network is repeatedly applied at a cost of increased operations to solve SISR problem. LapSRN \cite{LapSRN} aimed efficient super-resolution and used Laplacian pyramids to progressively extract features and reconstruct images at different scales with the same network. EDSR \cite{EDSR} improved the reconstruction results by eliminating Batch Normalization layers from the network and increasing the number of parameters to 43M. WDSR \cite{WDSR} further increased the parameters of the model to 75M and improved the results of EDSR. RDN \cite{RDN} used DenseNet \cite{DenseNet} style intermediate feature aggregation with residual blocks. More recently, researchers incorporated new ideas (such as grouped convolutions, attention layers etc.) into super resolution networks \cite{HAN, XLSR, GFIDN}. One obvious thing that can be deduced from these advancements is that as the number of parameters increases, the performance of the model increases as well. However, this comes at the cost of the model being practically not applicable. For these reasons, research interest in SISR has recently shifted towards building efficient models \cite{AIM2020,NTIRE22}. IDN \cite{IDN} follows this spirit; it uses channel splitting to distil features efficiently. IMDN further improves on this idea and uses channel splitting at granularity level and proposes information multi-distilling block (IMDB) which also includes a contrast-aware channel attention (CCA) layer. At the global level distilled information from the IMDBs are aggregated using Intermediate Information Collection (IIC). In this type of information collection, the information from the intermediate levels directly flows to the ends of the model. Indeed, this can be seen as a subset of the information collection type used in DenseNet and RDN, where DenseNet structure in RDN allows intermediate-to-intermediate flow as well. 

The problem of a deep learning model not being practically applicable is indeed a problem with other deep-learning models from different fields as well. Because of this, researchers proposed different approaches that can make a model run in real-time, such as, Hand Picked Architectures / Blocks, Network Pruning/ Sparsification, Knowledge Distilling, Quantization, Network Architecture Search (NAS)

Hand-picked architectures focuses on manually designed architectures and blocks. Network sparsification and pruning, such as \cite{MegviiPrune}, follow a different approach and try to eliminate the redundancies in a larger network to come up with a more efficient network. Knowledge distilling \cite{KnowDistill} uses heavy teacher and lighter student networks in a setting where the teacher network guides the student network. Quantization, such as \cite{QualcommQuant}, focuses on the deployment side and tries to keep the network performance under lighter arithmetic operations. Network architecture search \cite{NASNet} goes beyond these ideas and tries to find the network architecture in an optimization setting.

Indeed, these ideas can be used to design super-resolution networks as well. For this purpose, Li \etal \cite{DHP} proposed a differentiable pruning model. Their method reduced the number of parameters, FLOPS, and run-time of EDSR Baseline \cite{EDSR} and several other networks by a significant amount. In \cite{LWDNA}, Li \etal proposed Layer-Wise Differentiable Network Architectures to adjust the channel sizes of predefined networks and successfully reduced the number of parameters of EDSR Baseline while improving its performance. Song \etal \cite{NASSR} proposed an evolutionary network search algorithm for efficiently searching residual dense blocks for super-resolution networks. Wu \etal \cite{TriNAS} proposed a trilevel NAS algorithm for optimizing networks, cells, and kernels of super-resolution networks at the same time. In \cite{LowRankCNN} Li \etal followed a different approach for reducing the number of parameters, and proposed a learning basis for convolutional layers. Their method compresses the number of parameters of EDSR Baseline by up to 93\%.

While designing IMDeception we followed a manual approach since other approaches can still be applied to further push its limits.

\section{Proposed Method}

\begin{figure*}[htp]
\begin{subfigure}{.99\textwidth}
  \centering
  \includegraphics[scale=1]{"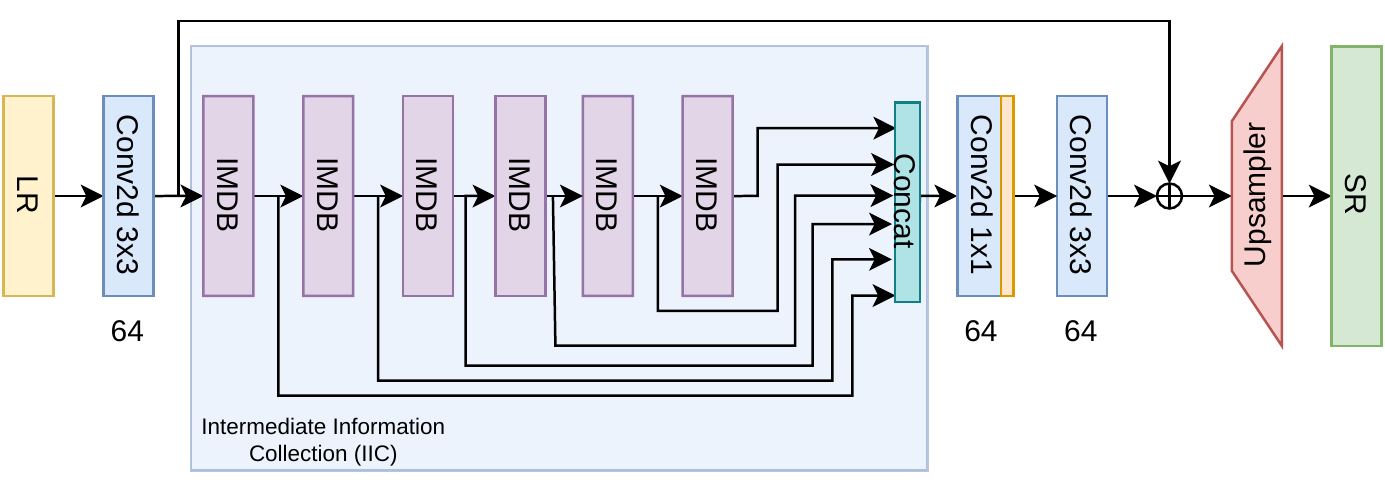"}
  \caption{IMDN Original Network: Note that here conv2d 1x1 includes a leaky relu activation.}
  \label{fig:IMDNNetwork}
\end{subfigure}%

\begin{subfigure}{.25\textwidth}
  \centering
  \vspace*{+0.7in}
  \includegraphics[scale=1]{"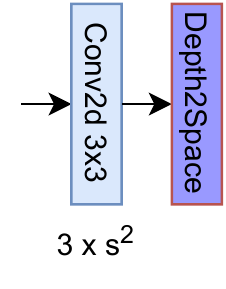"}
  \vspace*{+0.3in}
  \caption{Upsampler}
  \label{fig:Upsampler}
\end{subfigure}
\begin{subfigure}{.65\textwidth}
  \centering
  \vspace*{+0.3in}
  \includegraphics[scale=1]{"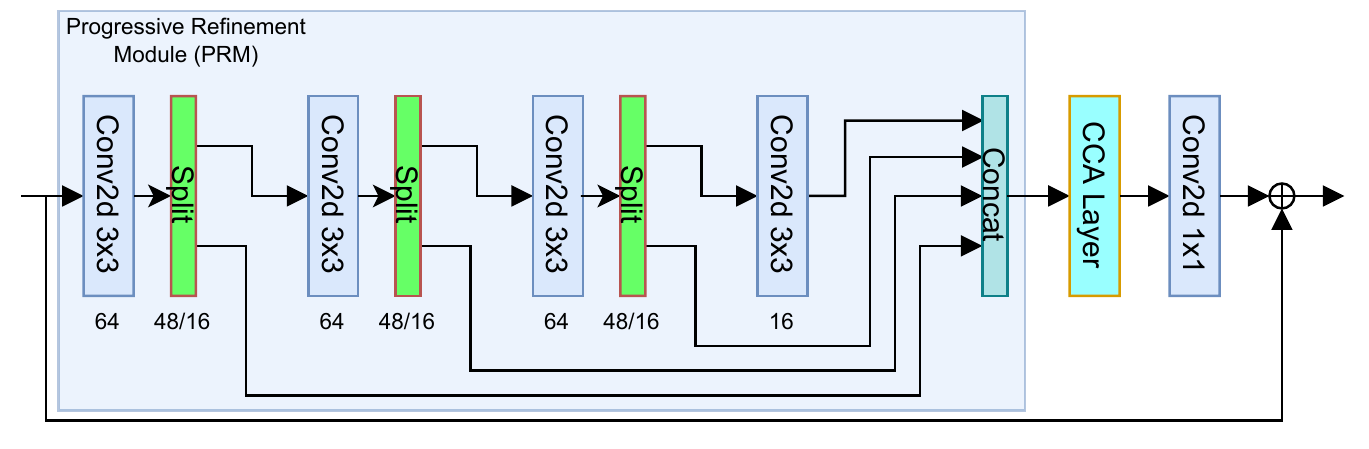"}
  \caption{IMDB Block: Here each conv2d 3x3 block includes a leaky relu activation}
  \label{fig:IMDBOrg}
\end{subfigure}
\caption{IMDN Structure and the submodules PRM and IIC are included here for reference. CCA layer details are omitted and can be found in the original work.}
\label{fig:IMDNStructure}
\end{figure*}

In this section, we describe the details of the proposed network. As it was mentioned before, the main motivation of this paper is efficiency while keeping performance at a comparable level with million-parameter networks. As a starting point, IMDN \cite{IMDN} is selected as the baseline of our work. The original IMDN architecture can be seen in \cref{fig:IMDNNetwork}.

\begin{figure*}[htp]
\begin{subfigure}{.99\textwidth}
  \centering
  \includegraphics[width=.99\textwidth]{"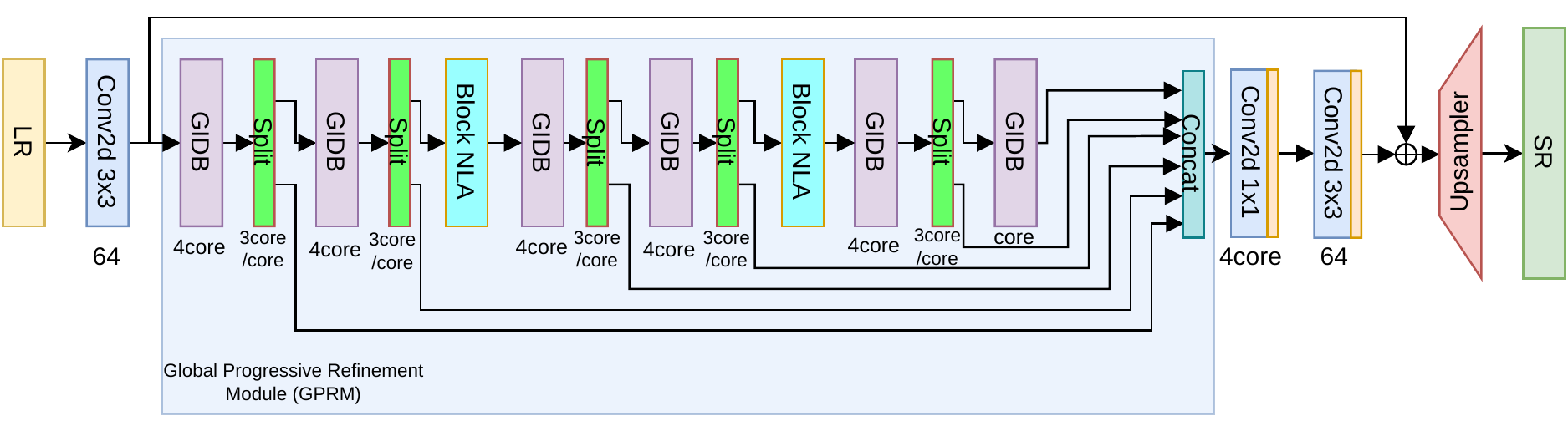"}
  \caption{IMDeception Network}
  \label{fig:IMDeceptionNetwork}
\end{subfigure}%
\vspace*{+0.3in}
\begin{subfigure}{.35\textwidth}
  \centering
  \includegraphics[width=.80\textwidth]{"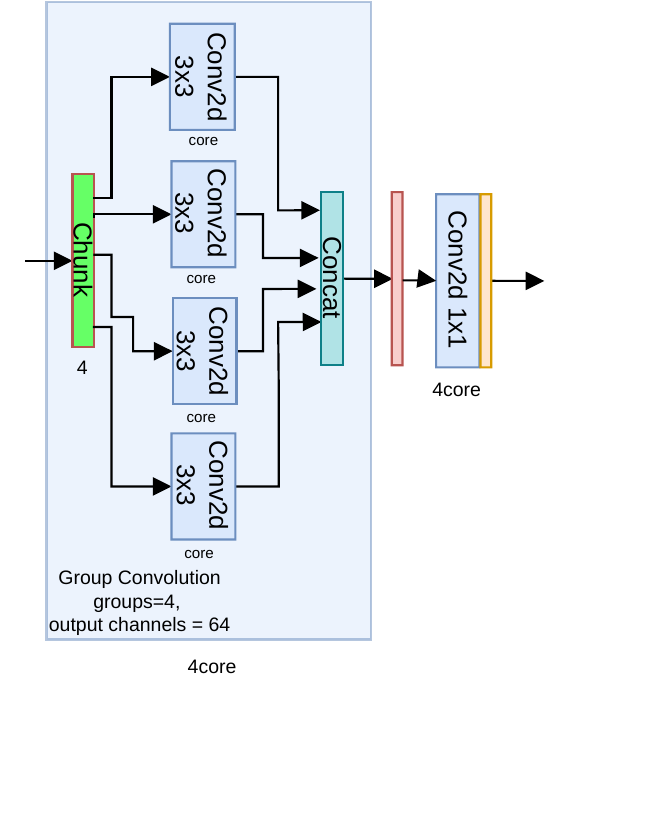"}
  \caption{Gblock: Here Red and Orange Stripes show ReLU and LeakyReLU activations respectively.}
  \label{fig:Gblock}
\end{subfigure}
\begin{subfigure}{.65\textwidth}
  \centering
  \vspace*{-0.1in}
  \includegraphics[width=.99\textwidth]{"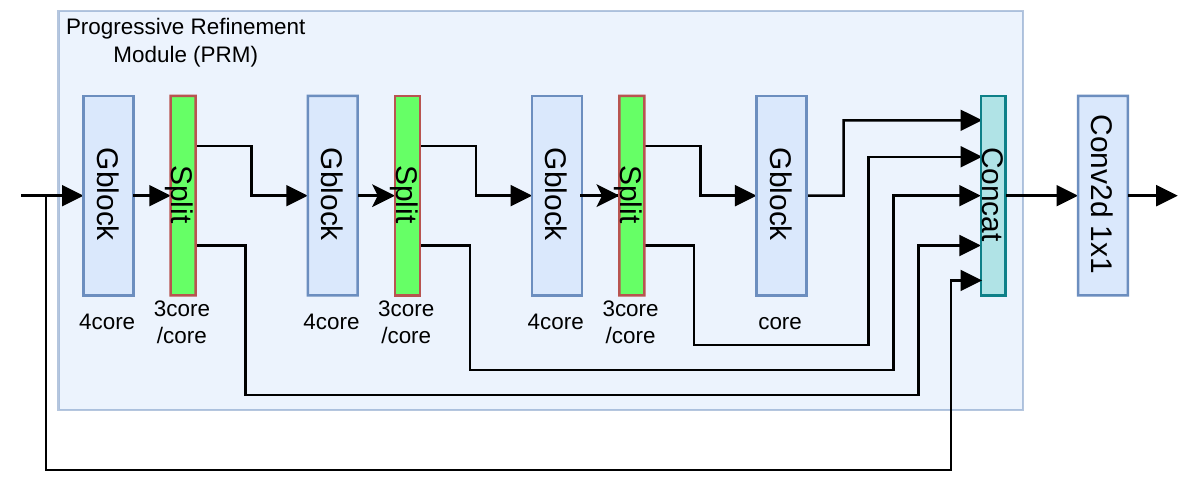"}
  \caption{GIDB Block}
  \label{fig:GIDB}
\end{subfigure}
\caption{IMDeception: Proposed Architecture}
\label{fig:IMDeception}
\end{figure*}

The variations of this network are already known to be high-performing \cite{NTIRE22,AIM2020,RFDN} and improving it is challenging. This is due to the already employed network mechanisms such as Progressive Refinement Module (PRM) (\cref{fig:IMDBOrg}) Contrast Aware Channel Attention (CCA) (\cref{fig:IMDBOrg}) and Intermediate Information Collection(IIC) (\cref{fig:IMDNNetwork}) are very efficient. The modules are indeed studied further in the original work, and each of these modules' contributions to the final model is noted. The individual contributions of each module can be seen in \cref{tab:deltaPRM}

\begin{table}[htp]
\centering
\begin{tabular}{||c c c c||} 
 \hline
 Module-Dataset & Set5 & Set14 & Manga109 \\ [0.5ex] 
 \hline\hline
 Basic & 31.86 & 28.43 & 29.92 \\ 
 \hline
  PRM Improvement & 0.15 & 0.06 & 0.24 \\ 
 \hline
 CCA Improvement & 0.09 & 0.02 & 0.09 \\
 \hline
 IIC Improvement & 0.01 & 0.01 & 0.03 \\ [1ex] 
 \hline
\end{tabular}
\caption{PRM, CCA and IIC modules' contributions to the basic network, *The table is derived from Table 3 of \cite{IMDN}}
\label{tab:deltaPRM}
\end{table}

Note that in IMDN, PRM and CCA are used locally inside the IMDBs, however IIC is used in a more global setting. Also note that the improvement provided by the PRM is much larger than the CCA and IIC. Furthermore the number of parameters drops with the PRM. Motivated by these facts and inspired by the Inception network's repeated structure \cite{Inception} we created a network structure where PRM is repeated locally in the blocks and globally among the blocks to improve performance and reduce the number of parameters. This is done in such a way that IIC in the global setting is replaced with the proposed Global PRM (\cref{fig:IMDeceptionNetwork}). Furthermore, CCA layers are used in every IMDB, but the performance contributions from these layers are marginal compared to the number of operations and parameters that they add to the network. However, since attention layers are great at increasing the receptive field, we decided to use a limited number of block-based non-local attention blocks \cite{NLA} in our proposed network's main path. To further reduce the number of parameters and number of operations of the network, every single Conv2D operation inside the IMDB is replaced with Gblocks (\cref{fig:Gblock}) as in XLSR \cite{XLSR} which is based on grouped convolutions. We call these group convolution based structures as Grouped Information Distilling Blocks (GIDB). Although the grouped convolutions are not well optimized in training frameworks, \cite{Gconv} if utilised correctly within an inference-oriented framework, group convolutions can lead to speed ups as noted in \cite{Gconv,XLSR} especially in mobile devices where efficient network structures are usually employed. 

Mathematically, the model can be described as follows; Given a low resolution image $I^{lr}$, super-resolved image, $I^{sr}$, can be obtained as follows:
\vspace{-0.0in}
\begin{gather}
    I^{sr} = H_{IMDeception}(I^{lr})
\label{mat:HIMDeception}
\end{gather}

Here, $H_{IMDeception}()$ is our proposed optimized super-resolution model. In the begining of the network a 64 channel 3x3 convolution is employed for feature extraction, as in IMDN, let $I^{feat}_{0}$ represent these features. These features are both transferred to the end of the network and processed in the Global Progressive Refinement Module as follows:
\vspace{-0.0in}
\begin{gather}
    I^{dfeat}_{1}, I^{sfeat}_{1} = Split(GIDB_{1}(I^{feat}_{0})) \nonumber\\
    I^{dfeat}_{2}, I^{sfeat}_{2} = Split(GIDB_{2}(I^{dfeat}_{1})) \nonumber\\
    I^{dfeat}_{2} = NLA(I^{dfeat}_{2}) \nonumber\\
    I^{dfeat}_{3}, I^{sfeat}_{3} = Split(GIDB_{3}(I^{dfeat}_{2})) \nonumber\\
    I^{dfeat}_{4}, I^{sfeat}_{4} = Split(GIDB_{4}(I^{dfeat}_{3})) \\
    I^{dfeat}_{4} = NLA(I^{dfeat}_{4}) \nonumber\\
    I^{dfeat}_{5}, I^{sfeat}_{5} = Split(GIDB_{5}(I^{dfeat}_{4})) \nonumber\\
    I^{sfeat}_{6} = GIDB_{6}(I^{dfeat}_{5}) \nonumber\\
    I^{feat}_{7} = Concat([I^{sfeat}_{1},...,I^{sfeat}_{6}]) \nonumber
\label{mat:GPRM}
\end{gather}

In the above equations, $Split(), Concat(), NLA()$ represent 3/1 ratio channel splitting, channel concatenation, and block based non-local attention. $I^{dfeat}_{n}, I^{sfeat}_{n}$ are channel split features of $n^{th}$ $GIDB$ block which is our proposed Grouped Information Distilling Block (GIDB). Note that here GPRM is used for global feature distilling and aggregation, and operating on the outputs of GIDBs. At the local level, features are further processed by GIDBs as follows:
\vspace{-0.1in}
\begin{gather}
    I^{dfeat1}_{n-1}, I^{sfeat1}_{n-1} = Split(GBlock_{1}(I^{dfeat}_{n-1})) \nonumber\\
    I^{dfeat2}_{n-1}, I^{sfeat2}_{n-1} = Split(GBlock_{2}(I^{dfeat1}_{n-1})) \nonumber\\
    I^{dfeat3}_{n-1}, I^{sfeat3}_{n-1} = Split(GBlock_{3}(I^{dfeat2}_{n-1})) \\
    I^{dfeat4}_{n-1}, I^{sfeat4}_{n-1} = Split(GBlock_{4}(I^{dfeat3}_{n-1})) \nonumber\\
    I^{dfeat}_{n} = H_{fuse}(Concat([I^{sfeat1}_{n-1},...,I^{sfeat4}_{n-1},I^{dfeat}_{n-1}]))\nonumber
\label{mat:GIDB}
\end{gather}

Here $H_{fuse}$ represents 1x1 convolution operation used for information fusion, Note that at the local level input features $I^{dfeat}_{n-1}$ are processed and refined in grouped fashion using $Gblock()$s. $Gblock()$ is implemented using 3x3 grouped convolution (groups=4) and cascaded 1x1 convolution to allow information flow between the groups. Using information grouping and processing the features in a grouped fashion reduces the number of parameters while almost at no cost of performance loss. The detailed $Gblock()$ implementation along with the used activation functions can be seen in \cref{fig:Gblock}.

The output of the GPRM module, $I^{feat}_{7}$, is further processed to construct the super-resolved image, $I^{sr}$ as follows:
\vspace{-0.1in}
\begin{gather}
    I^{feat}_{8} = LeakyReLU(H_{1x1}(I^{feat}_{7})) \nonumber\\
    I^{feat}_{9} = LeakyReLU(H_{3x3}(I^{feat}_{8})) + I^{feat}_{0} \\
    I^{sr} = Upsample(I^{feat}_{9}) \nonumber
\label{mat:LastProcess}
\end{gather}

Here $H_{1x1},H_{3x3}$ represent 1x1 and 3x3 convolution layers respectively. We used Leaky ReLU (slope=0.05) activations. $Upsample()$ is the upsampling layer implemented as shown in \cref{fig:Upsampler}

Our proposed network structure, which we call IMDeception, combining all of these ideas, can be seen in \cref{fig:IMDeception}.

Note that we used global PRM among the GIDB and local PRM as in IMDB inside GIDB. Our proposed architecture defines a class of highly efficient architectures sharing the same structure with different channel numbers on the filters. As it can be seen from \cref{fig:IMDeception}, we define the complexity of the models using $core$ parameter. Depending on the needs, the $core$ parameter can be used to adjust the complexity of the network. From our experiments, we have observed that even $core=4$ with no attention blocks can still show high reconstruction performance with great inference timings.

The performance parameters of various IMDeception networks using $core$ parameter and existence of attention blocks can be seen in \cref{tab:perfparams}

\begin{table*}[htp]
\begin{center}
\begin{tabular}{||c | c | c | c  | c | c ||} 
 \hline
  IMDeception & $core=16$ + NLA & $core=12$ & $core=8$ &  $core=4$ + NLA&  $core=4$ \\
 \hline\hline
 \#Params & $316K$ & $198K$ & $113K$ & $57K$ & $57K$ \\ 
 \hline
  FLOPS[G] & $20.7$ & $12.9$ & $7.4$ & $3.7$ & $3.7$ \\ 
 \hline
 Act.[M] & $206$ & $149$ & $103$ & $57$ & $57$ \\ 
 \hline
  Runtime[ms]$^*$ & 60 & 45 & 31 & 24 & 21 \\ 
  \hline
  \#Conv2d & 62 & 58 & 58 & 62 & 58 \\ 
 \hline
  \#Div2K Val. (PSNR) & 29.02 & 28.82 & 28.70 & 28.48 & 28.45 \\ 
 \hline
\end{tabular}

\caption{IMDeception performance parameters, \\ *Averaged on Div2K Validation Set on NVIDIA 2080 Super}
\label{tab:perfparams}
\end{center}
\end{table*}

\section{Experiments}

\subsection{Datasets}
For the training, we used Div2K Dataset \cite{Div2K}, and Flickr2K \cite{EDSR} combined (DF2K). In total, the training set includes, 3450 images, and for validation we used Div2K validation set, which includes 100 images.  

\subsection{Training Details}

The proposed model ($core=16+NLA$) was trained in two different phases in all of the phases we used;

\begin{figure}[htp]
\begin{subfigure}{.5\textwidth}
  \centering
  \includegraphics[width=.99\textwidth]{"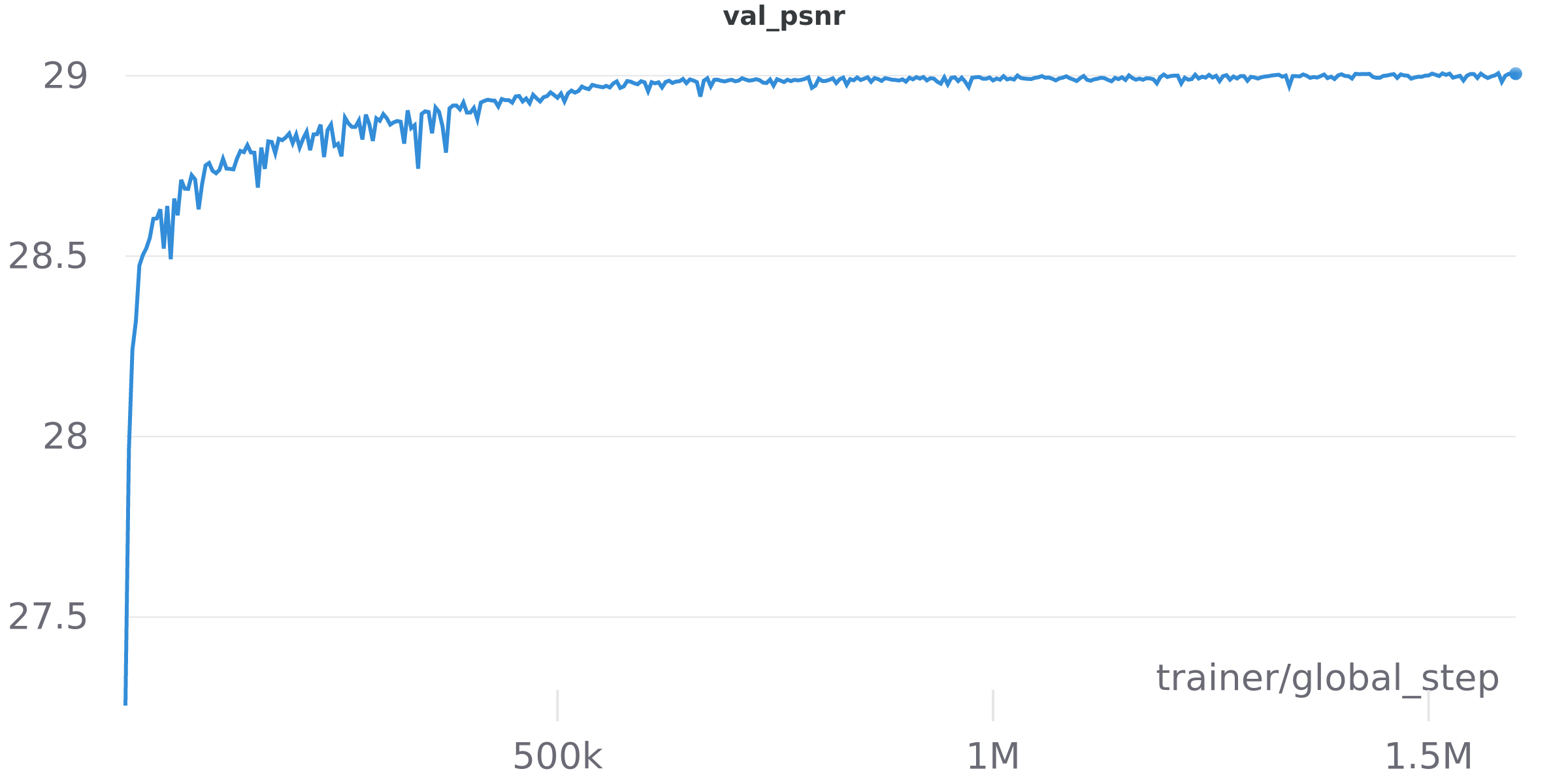"}
  \caption{Validation PSNR of the 1st Phase of Training}
  \label{fig:val_psnr}
\end{subfigure}
\begin{subfigure}{.5\textwidth}
  \centering
  \includegraphics[width=.99\textwidth]{"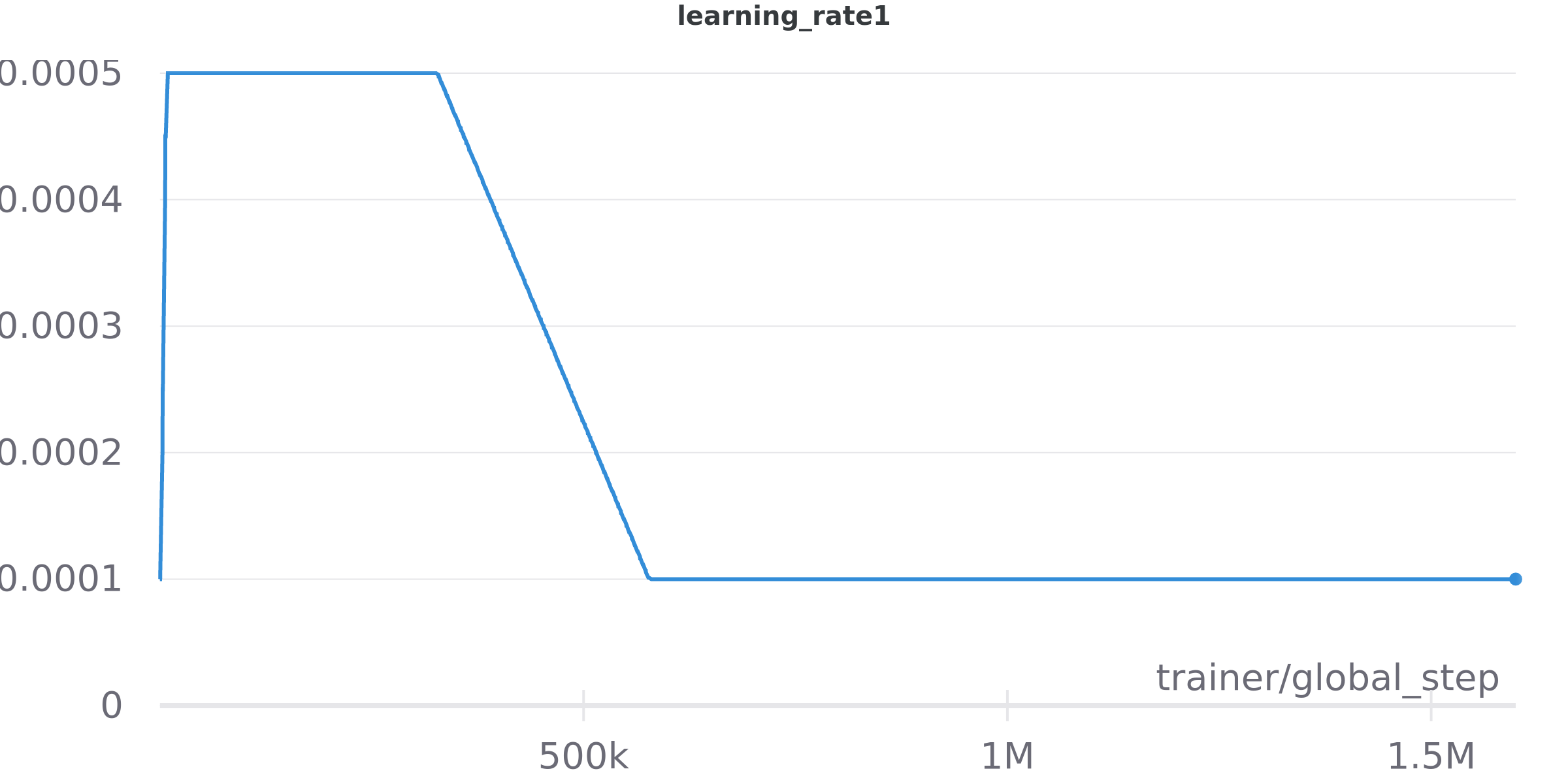"}
  \caption{Knee Learning Rate Scheduler Policy used for the training phases}
  \label{fig:LR}
\end{subfigure}
\caption{Training Curves}
\label{fig:TrainingCurves}
\end{figure}

\begin{itemize}
\setlength\itemsep{0.01em}
\item Adam optimizer with $\beta_1=0.9, \beta_2=0.999$.
\item Mini-batch size of 8.
\item Cropped HR Image Size of 512 x 512.
\item Zero padding is used when necessary.
\item Each epoch contains 800 mini batches.
\item Knee Learning Rate Scheduling \cite{KneeLR} with 10 epochs warm-up 400 epochs exploit and 400 epochs cool-down period with maximum learning rate $0.5e^{-3}$ (\cref{fig:LR}).
\item PyTorch model is trained within PyTorch-Lightning framework.
\end{itemize}

For the first phase, we used Charbonnier loss with $\epsilon=0.1$ as in \cref{mat:charbon} and trained for 2000 epochs, which lasted 2 days and 7 hours on a single NVIDIA Tesla v100. See \cref{fig:TrainingCurves} for training curves and learning rate policy.
\vspace*{-0.05in}
\begin{gather}
    Charbonnier(x) = \sqrt{x^2+\epsilon^2}
\label{mat:charbon}
\end{gather}

The second phase of the training started from the best checkpoint, and this time L2-norm was used as the loss function, and trained for 1300 epochs which lasted 1 day and 16 hours.

\subsection{Results}

\begin{figure}[htp]
\centering
\includegraphics[width=.5\textwidth]{"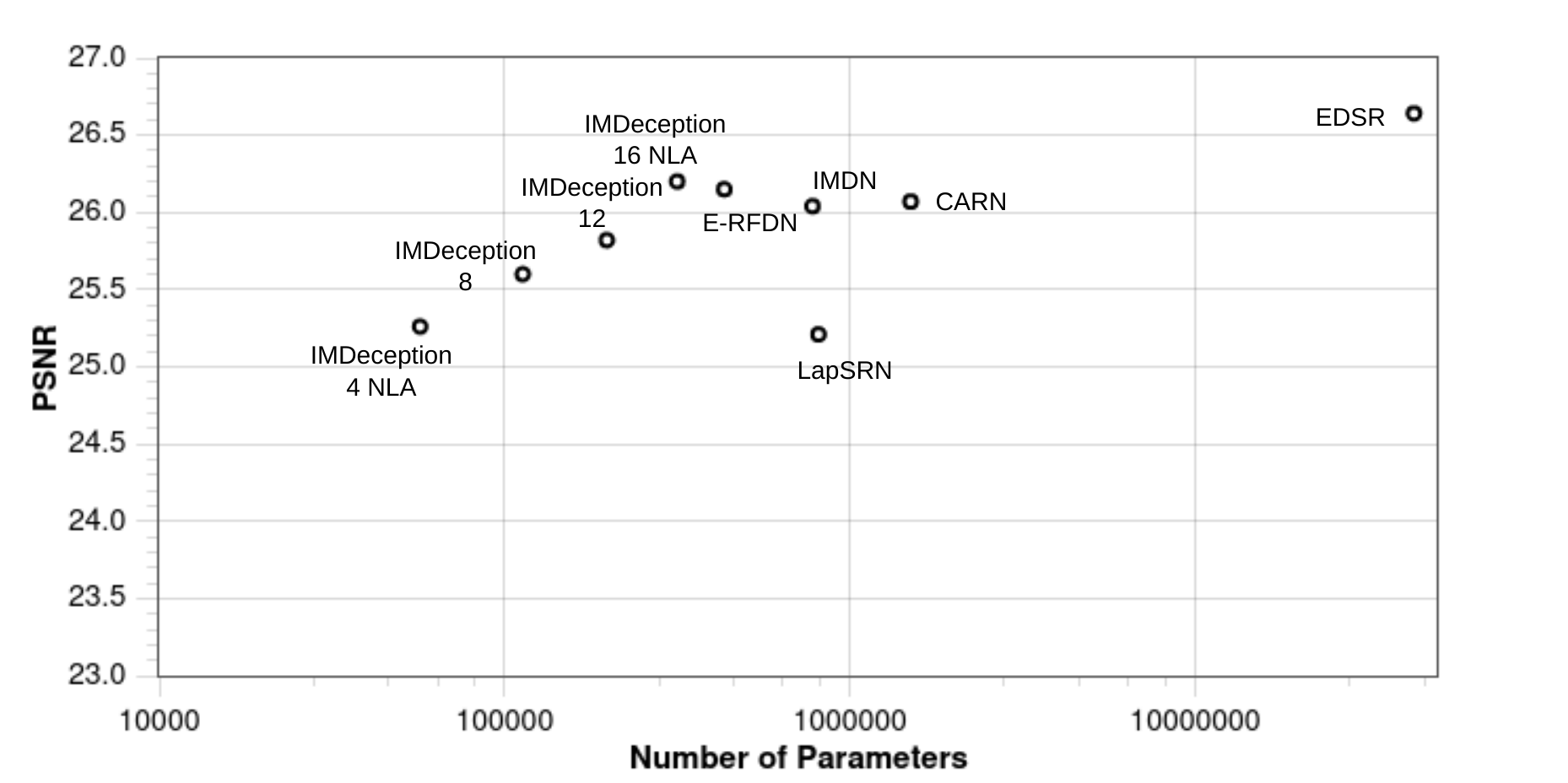"}
\caption{PSNR vs Number of Parameters for Urban100 Dataset for Different Super-Resolution Models}
\label{fig:ParamsPSNR}
\end{figure}

In this section, the proposed architecture's PSNR results are given on various different datasets. The PSNR results of IMDeception and other state-of-the-art methods can be seen in \cref{tab:PSNR}. From the experiments it can be seen that, although IMDeception ($core=16+NLA$) has very limited number of parameters and FLOPS, it has on par performance with state-of-the-art algorithms. Especially, IMDeception's performance on Urban100, Manga109 datasets is well above E-RFDN, IMDN, CARN, LapSRN methods except EDSR (which has 43M parameters). An interesting result is IMDeception ($core=4$)'s PSNR performance on Urban100 and Manga109 surpasses LapSRN although it has only 7\% of number of parameters. The number-of-parameters and PSNR results of these methods can be best seen in \cref{fig:ParamsPSNR}

\begin{figure*}
\centering

\begin{tabular}{ccc}
\begin{tabular}{@{}c@{}} 
\subfloat[Original: BSD100 8023.png]{\includegraphics[width=0.45\linewidth]{"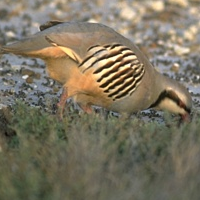"}} \\ 
\end{tabular} & 
\begin{tabular}{@{}c@{}} 
\subfloat[Bicubic]{\includegraphics[trim={90 70 0 20},clip,width=0.22\linewidth]{"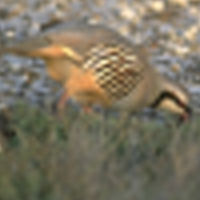"}} \\ 
\subfloat[CARN]{\includegraphics[trim={90 70 0 20},clip,width=0.22\linewidth]{"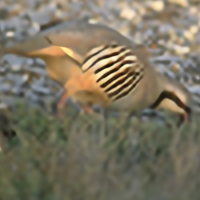"}} 
\end{tabular} & 
\begin{tabular}{@{}c@{}} 
\subfloat[\textbf{IMDeception (ours)}]{\includegraphics[trim={90 70 0 20},clip,width=0.22\linewidth]{"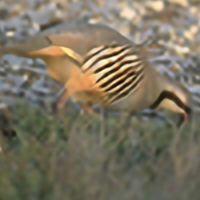"}} \\ 
\subfloat[IMDN]{\includegraphics[trim={90 70 0 20},clip,width=0.22\linewidth]{"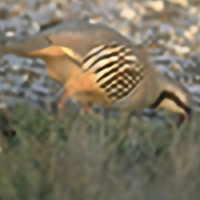"}} 
\end{tabular}
\end{tabular}

\vspace{5mm}

\begin{tabular}{ccc}
\begin{tabular}{@{}c@{}} 
\subfloat[Original: Urban100 img\_048.png]{\includegraphics[width=0.45\linewidth]{"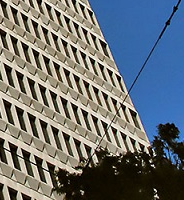"}} \\ 
\end{tabular} & 
\begin{tabular}{@{}c@{}} 
\subfloat[Bicubic]{\includegraphics[width=0.22\linewidth]{"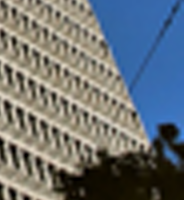"}} \\ 
\subfloat[CARN]{\includegraphics[width=0.22\linewidth]{"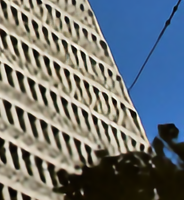"}} 
\end{tabular} & 
\begin{tabular}{@{}c@{}} 
\subfloat[\textbf{IMDeception (ours)}]{\includegraphics[width=0.22\linewidth]{"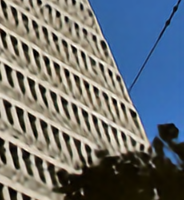"}} \\ 
\subfloat[IMDN]{\includegraphics[width=0.22\linewidth]{"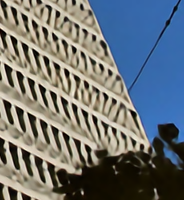"}} 
\end{tabular}
\end{tabular}

   \caption{Example Images from Reference Datasets. IMDeception is compared with other methods}
   
\label{fig:Div2kcompare}
\end{figure*}

Another important property that IMDeception has, its precise output on the repeated structures and patterns which can be seen in \cref{fig:Div2kcompare}. 

\begin{table*}
\centering

\begin{tabular}{|l|l|l|l|l|l||l|}
\hline
Model                                                               & Set5  & Set14 & BSD100 & Urban100 & Manga109 & Div2K (Val) \\ \hline
Bicubic                                                             & 28.42 & 26.00 & 25.96  & 23.14    & 24.89    & 26.66       \\ \hline
LapSRN (812K)                                                             & 31.54 & 28.19 & 27.32  & 25.21    & 29.09    & 28.75       \\ \hline
IMDN  (779K)                                                              & 32.21 & 28.58 & 27.56  & 26.04    & 30.45    & 28.94       \\ \hline
EDSR  (43M)                                                              & 32.46 & 28.80 & 27.71  & 26.64    & 31.02    & 29.25       \\ \hline
CARN\cite{CARN} (1.5M)                                                              & 32.14 & 28.61 & 27.58  & 26.07    & 30.46    & 28.96       \\ \hline
E-RFDN  (433K)                                                            & 32.16 & 28.65 & 27.60  & 26.15    & 30.59    & 29.04           \\ \hline
\hline
\begin{tabular}[c]{@{}l@{}}\textbf{IMDeception} (316K)\\ core=16 + NLA\end{tabular} & 32.14 & 28.64 & 27.60  & 26.20    & 30.67    & 29.02$^*$       \\ \hline
\begin{tabular}[c]{@{}l@{}}\textbf{IMDeception}(198K)\\ core=12\end{tabular}       &  31.83     &  28.47     &  27.50      &  25.82        &  30.28    &  28.83           \\ \hline
\begin{tabular}[c]{@{}l@{}}\textbf{IMDeception}(113K)\\ core=8\end{tabular}        &  31.69     &  28.35      &  27.42      &  25.60        &  29.89  &   28.69        \\ \hline
\begin{tabular}[c]{@{}l@{}}\textbf{IMDeception}(57K)\\ core=4\end{tabular}        &   31.35    &  28.12     &  27.27      &   25.23       &   29.20       & 28.45            \\ \hline
\begin{tabular}[c]{@{}l@{}}\textbf{IMDeception}(57K)\\ core=4 + NLA\end{tabular}  &   31.42    &  28.18     &  27.26      &   25.26       &   29.28       & 28.48            \\ \hline
\end{tabular}

\caption{Experimental results of the proposed method were compared with various different methods for x4 scaling. Note that except for Div2K Validation result, all PSNR values are calculated on Luminance (Y) channel to be consistent with the literature. \\ $^*$ Div2K Test Set PSNR is 28.73}
\label{tab:PSNR}
\end{table*}

In terms of run-time, our proposed method has great potential for optimization on edge devices \cite{Gconv}, due to parallel grouped convolutions and a reduced number of parameters. As it can be seen from the \cref{tab:PSNR}, the proposed model defines a set of efficient architectures, which can be used in different devices with different inference run-times with good reconstruction performance. As a reference and as an indication of its potential, we run our proposed models on NVIDIA RTX 2080 Super and NVIDIA Jetson Xavier AGX 30W devices. To do this, we have converted trained models to ONNX format and used NVIDIA's TRT Engine application to convert them to an inference engine to use the hardware's full potential. The run-times are listed in \cref{tab:Inference}. Note that IMDeception can run on this edge device at up to 24fps while outputting high-resolution 2K images. An important conclusion that can be made from the run-time experiments although the number of parameters and FLOPS are lower for IMDeception $core=12$ compared to $core=16 + NLA$, the inference run-times are higher. This is due to the fact that GPUs are usually optimized for the channel sizes, which are powers of 2. Because 12 is not a power of 2, additional processing in the GPU is required, negating the benefits of the reduced number of parameters and FLOPS. This is an important conclusion to make since this phenomenon is not observable during inference with a training framework such as PyTorch.

\begin{table}[htp]
\centering

\begin{tabular}{|l|l|l|}
\hline
Model\textbackslash{}Hardware                                    & \begin{tabular}[c]{@{}l@{}} {\footnotesize RTX 2080} \\ {\footnotesize TensorRT (ms)}\end{tabular} & \begin{tabular}[c]{@{}l@{}}{\footnotesize Jetson Xavier} \\ {\footnotesize TensorRT (ms)}\end{tabular} \\ \hline
\begin{tabular}[c]{@{}l@{}}IMDeception\\ core=16+NLA\end{tabular}                                                          & 17.7                                                           & 88.9                                                                \\ \hline
\begin{tabular}[c]{@{}l@{}}IMDeception$^*$\\ core=12\end{tabular}                                                              & 19                                                           & 98.9                                                                \\ \hline
\begin{tabular}[c]{@{}l@{}}IMDeception\\ core=8\end{tabular}                                                               & 9.9                                                           & 51.1                                                                \\ \hline
\begin{tabular}[c]{@{}l@{}}IMDeception\\ core=4+NLA\end{tabular}                                                           & 9.8                                                           & 44.7                                                                \\ \hline
\begin{tabular}[c]{@{}l@{}}IMDeception\\ core=4\end{tabular}                                                               & 9.2                                                           & 41.9                                                                \\ \hline
\end{tabular}

\caption{Inference Results for an input image size of 512x256 \\ $^*$ Note the increased inference time. This is because 12 is not a power of 2 and the GPUs are optimized and have kernels specific to sizes of power of 2.}
\label{tab:Inference}
\end{table}

\section{Conclusion}

We proposed an efficient model based on the IMDN network called IMDeception. IMDeception employs the proposed Global Progressive Refinement Module (GPRM), which is an extension of the Progressive Refinement Module (PRM). Unlike PRM that works only with the Conv2d layer at the local scale, GPRM can be used with any arbitrary block, as we did with the newly proposed Grouped Information Distilling Blocks (GIDB). Both of the proposed mechanisms/blocks can be used for different networks and in different structures. These structures are designed with efficiency in mind, which reduces the number of parameters and FLOPS while maintaining high performance. GPRM is an efficient way of combining features and can be an alternative to Dense Networks style or IIC-style feature aggregating methods. One nice feature of it is that it separates the aggregated part from the distilled part, which helps controlling the network size while maintaining network performance. On the other hand, GIDB uses grouped convolutions, which, if implemented with efficiency in mind, can provide a speed boost during inference. We also showed that the proposed model is very high-performing on various different datasets and has great inference timings on different hardware, including NVIDIA Jetson Xavier AGX.

{
\small
\bibliographystyle{ieee_fullname}
\bibliography{egbib}
}

\end{document}